\documentclass[10pt,conference]{IEEEtran}
\IEEEoverridecommandlockouts
\usepackage{cite}
\usepackage{amsmath,amssymb,amsfonts}
\usepackage{algorithmic}
\usepackage{graphicx}
\usepackage{textcomp}
\usepackage{xcolor}
\def\BibTeX{{\rm B\kern-.05em{\sc i\kern-.025em b}\kern-.08em
    T\kern-.1667em\lower.7ex\hbox{E}\kern-.125emX}}
\begin{document}

\title{Training-Free Light-Guided Text-to-Image Diffusion Model via Initial Noise Manipulation}

% \author{
% \IEEEauthorblockN{Anonymous}

% }
\author{
Ryugo Morita$^{1,2}$, Stanislav Frolov$^{1}$, Brian Bernhard Moser$^{1}$, Ko Watanabe$^{1}$, Riku Takahashi$^{1,2}$, Andreas Dengel$^{1}$\\
$^{1}$RPTU Kaiserslautern-Landau \& DFKI GmbH, Kaiserslautern, Germany \\
$^{2}$Faculty of Science and Engineering, Hosei University, Tokyo, Japan

% {\tt\small ryugo.morita@dfki.de}
}

\maketitle

\begin{abstract}
Diffusion models have demonstrated high-quality performance in conditional text-to-image generation, particularly with structural cues such as edges, layouts, and depth. However, lighting conditions have received limited attention and remain difficult to control within the generative process.
Existing methods handle lighting through a two-stage pipeline that relights images after generation, which is inefficient. Moreover, they rely on fine-tuning with large datasets and heavy computation, limiting their adaptability to new models and tasks.
To address this, we propose a novel Training-Free Light-Guided Text-to-Image Diffusion Model via Initial Noise Manipulation (LGTM), which manipulates the initial latent noise of the diffusion process to guide image generation with text prompts and user-specified light directions. Through a channel-wise analysis of the latent space, we find that selectively manipulating latent channels enables fine-grained lighting control without fine-tuning or modifying the pre-trained model.
Extensive experiments show that our method surpasses prompt-based baselines in lighting consistency, while preserving image quality and text alignment. This approach introduces new possibilities for dynamic, user-guided light control. Furthermore, it integrates seamlessly with models like ControlNet, demonstrating adaptability across diverse scenarios.
\end{abstract}


\section{Introduction}
\label{sec:intro}
Diffusion models have revolutionized image synthesis by enabling high-fidelity text-to-image generation~\cite{song2020denoising,ho2020denoising,rombach2022high}, with wide-ranging applications such as art, design, and education~\cite{lee2025semanticdraw,wang2025designdiffusion,morita2025genaireading}. 
To better reflect user preferences, recent studies have explored conditional generation using structural cues such as edges, segmentation maps, and layouts~\cite{zhang2023adding,koley2024s,balaji2022ediff,gafni2022make,zheng2023layoutdiffusion, mao2023training}.
These methods focus on object features, with limited attention to lighting, an essential factor for realism and mood.

% -------------------------------------------------------------------------
\begin{figure}[t]
    \centering
    \includegraphics[width=\linewidth]{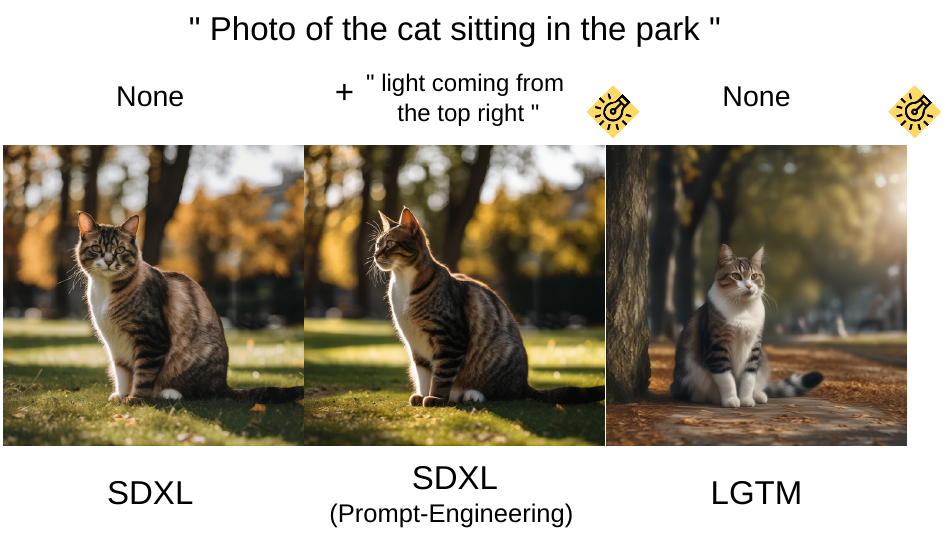} 
    \caption{Existing prompt-engineering methods fail to generate differences in generated images with or without light-specific prompts, resulting in outputs that overlook specified lighting conditions. Our proposed LGTM effectively guides lighting during image generation, ensuring outputs align with text prompts and desired lighting directions without fine-tuning.}
    \label{fig:teaser}
\end{figure}
% -------------------------------------------------------------------------

Recent works~\cite{zhang2025scaling, zeng2024dilightnet} address lighting control via two-stage workflows that first generate an image and then apply a separate relighting module to modify its illumination. However, such pipelines are inefficient and typically depend on illumination-annotated datasets and heavy fine-tuning. IC-Light~\cite{zhang2025scaling} is trained on approximately 10 million images using 8×H100 (80GB) GPUs over 100 hours, while DelightNET~\cite{zeng2024dilightnet} constructs a synthetic dataset of 25K objects, each rendered under 4 viewpoints and 12 lighting conditions, and is trained with 8×V100 GPUs for 30 hours. In the fast-evolving landscape of generative models, such resource-heavy pipelines are increasingly impractical.

On the other hand, prompt engineering~\cite{liu2022design} provides a lightweight, training-free way to influence generation, but it remains unreliable for controlling illumination. As shown in Fig.~\ref{fig:teaser}, Stable Diffusion~\cite{rombach2022high} fails to achieve consistent illumination even when users explicitly specify a light direction via text prompt. This highlights the need for a more direct and seamless method to incorporate user-defined lighting conditions into the generation process.

To address this challenge, we propose a novel Training-Free Light-Guided Text-to-Image Diffusion Model (LGTM) that manipulates the initial noise to steer illumination throughout the diffusion process. 
We first conduct a channel-wise sensitivity analysis of the VAE latent noise in Latent Diffusion Models (LDMs), and find that \emph{channel 1} is strongly correlated with global brightness and perceived light direction.
Guided by this analysis, LGTM selectively manipulates \emph{channel 1} to enable intuitive and fine-grained lighting control without fine-tuning.

Extensive experiments demonstrate that LGTM achieves more accurate and coherent lighting aligned with user-specified directions than prompt-based methods in Stable Diffusion, while preserving visual quality and text–image alignment.
In addition, by modifying only the initial noise, our method can be seamlessly applied to conditional modules such as ControlNet~\cite{zhang2023adding}, enabling simultaneous control over structural cues (e.g., edges) and illumination, and demonstrating strong adaptability to diverse generation scenarios and user constraints. Our contributions are as follows:
\setlength{\leftmargini}{1em} 
\begin{itemize}
    \item We define light-guided text-to-image generation as a novel task and propose a Training-Free Light-Guided Text-to-Image Diffusion Model (LGTM) to address this.
    \item We are the first to explore light control via latent-space manipulation by leveraging the disentangled structure of the VAE latent channels in LDM, identifying \emph{channel 1} as a key factor for encoding lighting information.
    \item LGTM achieves effective illumination control by modifying only the initial latent noise, without altering the model architecture or parameters, making it compatible with extended frameworks like ControlNet and adaptable to a wide range of image/video generation scenarios.
\end{itemize}

\section{Related Works}
\label{sec:related}

\subsection{Conditional Text-to-Image Generation}
Diffusion models have significantly advanced text-to-image generation~\cite{saharia2022photorealistic, rombach2022high}. 
Conditional text-to-image methods incorporate additional modalities, such as edges~\cite{zhang2023adding, koley2024s}, segmentation~\cite{balaji2022ediff, gafni2022make}, and layouts~\cite{zheng2023layoutdiffusion,yang2023reco}, to better align generated images with user preferences.
As image generation models rapidly evolve, training-based approaches suffer from limited adaptability due to the need for extensive retraining.
This has motivated growing interest in training-free conditional generation methods. Recent training-free approaches mainly manipulate attention mechanisms to control structural properties, such as object layouts~\cite{mao2023training, mo2024freecontrol}, inter-object relationships~\cite{chefer2023attend,li2025countdiffusion}, and panoramic compositions~\cite{bar2023multidiffusion,frolov2025spotdiffusion}.
However, lighting remains unexplored in training-free settings.
Unlike structural attributes, illumination requires costly annotations or curated datasets with diverse lighting conditions.

\subsection{Initial Noise of Diffusion Model}
Diffusion models begin with Gaussian noise and denoise it to synthesize images aligned with a target distribution. The initial noise plays a pivotal role in this process, as they strongly influence both visual quality and semantic alignment~\cite{yan2025beyond,li2024all,qi2024not}. 
Recent studies optimize the initial noise with human preference signals or attention-based metrics to improve image fidelity and text-image alignment without requiring additional model training~\cite{sundaram2025cocono,eyring2024reno}. Furthermore, selecting optimal noise seeds or modifying localized regions within the noise affects object placement and overall image quality~\cite{eyring2025noise, ban2024crystal}.
Beyond optimization, direct manipulation of the initial noise is leveraged to exert fine-grained control over image layouts, enabling layout-aware and layered image generation~\cite{mao2024theLottery,morita2025sawna,morita2024tkg,nagai2025taue}. Inspired by these findings, we explore illumination control beyond structural control via initial noise manipulation.

\subsection{Diffusion Models for Light Control}
Existing lighting control methods primarily focus on relighting existing images rather than incorporating lighting guidance directly into the text-to-image generation process.
Relightful Harmonization~\cite{ren2024relightful} adjusts illumination through post-processing, while other approaches such as DelightNET~\cite{zeng2024dilightnet}, FlashTex~\cite{deng2025flashtex}, and LightIt~\cite{kocsis2024lightit} rely on 3D rendering or synthetic datasets.
IC-Light~\cite{zhang2025scaling} further introduces a physically motivated light transport mechanism during training.

Despite their effectiveness, these methods rely on a two-stage relighting paradigm that modifies illumination after image generation.
In contrast, we define light-guided text-to-image generation as a new task, where lighting conditions are specified at generation time.
We address this task with a training-free approach that directly integrates lighting control into the diffusion process, without a separate relighting stage.

% -------------------------------------------------------------------------
\begin{figure*}[t]
    \centering
    \includegraphics[width=0.9\linewidth]{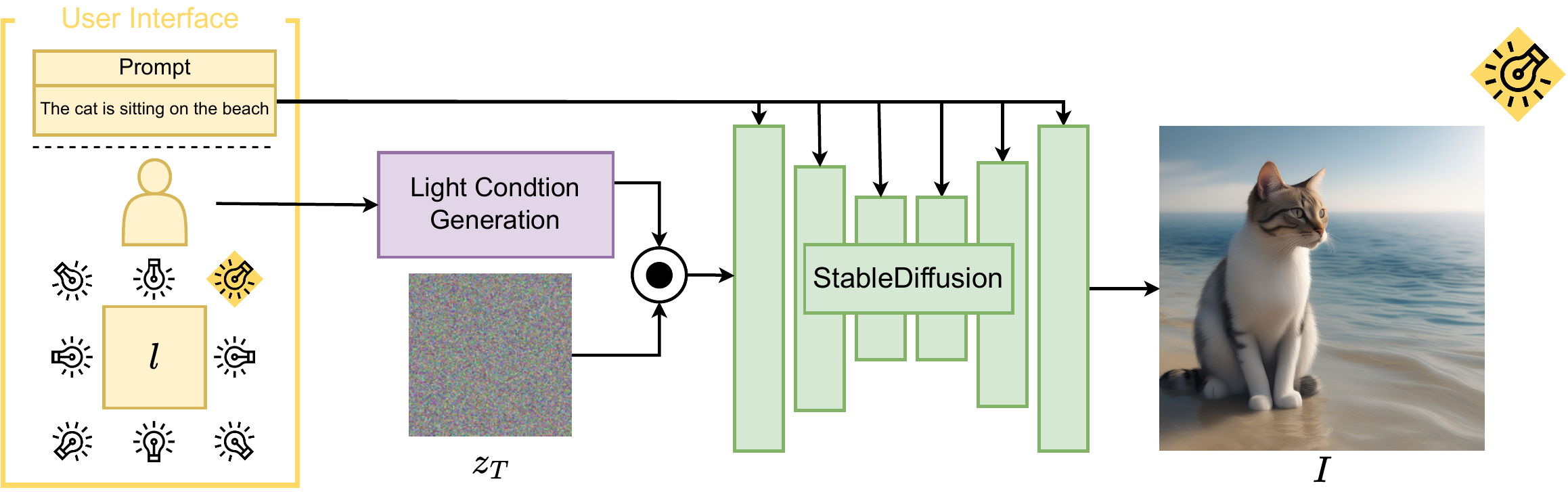}
    \caption{Overview of our proposed LGTM. The user inputs a prompt $p$ and a light condition $l$. The Light Conditional Generation module generates the light direction mask $m_l$ according to $l$ for manipulating the initial noise in Stable Diffusion. Then, the vanilla Stable Diffusion model integrates these inputs, dynamically adjusting the latent space—particularly \emph{channel 1}—to reflect user-defined lighting conditions. Finally, it outputs the final image $I$.}
    \label{fig:model}
\end{figure*}
% -------------------------------------------------------------------------

\section{Methods}
\label{sec:methods}
As illustrated in Fig.~\ref{fig:model}, LGTM takes a text prompt $p$ and a user-specified light direction $l$ as inputs, and generates an output image $I$ whose illumination follows $l$ while remaining consistent with $p$.
To achieve this, we first analyze channel-wise sensitivities of the initial latent noise in the VAE latent space, and then guide illumination by manipulating the initial noise according to a light mask derived from $l$.

\subsection{Preliminaries}
\label{sec:preliminaries}
Our method extends Latent Diffusion Models (LDM)~\cite{rombach2022high}, operating in the latent space of a VAE encoders $\mathcal{E}$ and decoders $\mathcal{D}$~\cite{kingma2013auto}. The encoder $\mathcal{E}$ encodes an image $x \in \mathbb{R}^{H \times W \times 3}$ in RGB space into a latent representation $z = \mathcal{E}(x)$, where $z \in \mathbb{R}^{H/8 \times W/8 \times 4}$ and the decoder $\mathcal{D}$ is trained to reconstruct $x$ as $\hat{x} = \mathcal{D}(z)$, which is approximately identical to $x$. 

Stable Diffusion employs a Denoising Diffusion Probabilistic Model (DDPM)~\cite{ho2020denoising} operating in the latent space of LDM. It trains a U-Net model, $\epsilon_\theta$, to predict noise added to an initial latent, denoted as $z_t$, which is the latent $z = \mathcal{E}(x)$ with noise added at timestep $t \in T$. Given a condition $y$ (i.e., text prompt), the objective of a text-to-image LDM is
\begin{equation}
    \mathcal{L}_{\text{LDM}} := \mathbb{E}_{\mathcal{E}(x), y, \epsilon \sim \mathcal{N}(0,1), t}\left[\|\epsilon - \epsilon_\theta(z_t, t, \tau_\theta(y))\|^2\right],
\end{equation}
where both $\epsilon_\theta$ and $\tau_\theta$ are jointly optimized. However, Stable Diffusion adopts a frozen CLIP text encoder instead of a trainable text encoder $\tau_\theta$. 

% -------------------------------------------------------------------------
\begin{figure}[t]
    \centering
    \includegraphics[width=\linewidth]{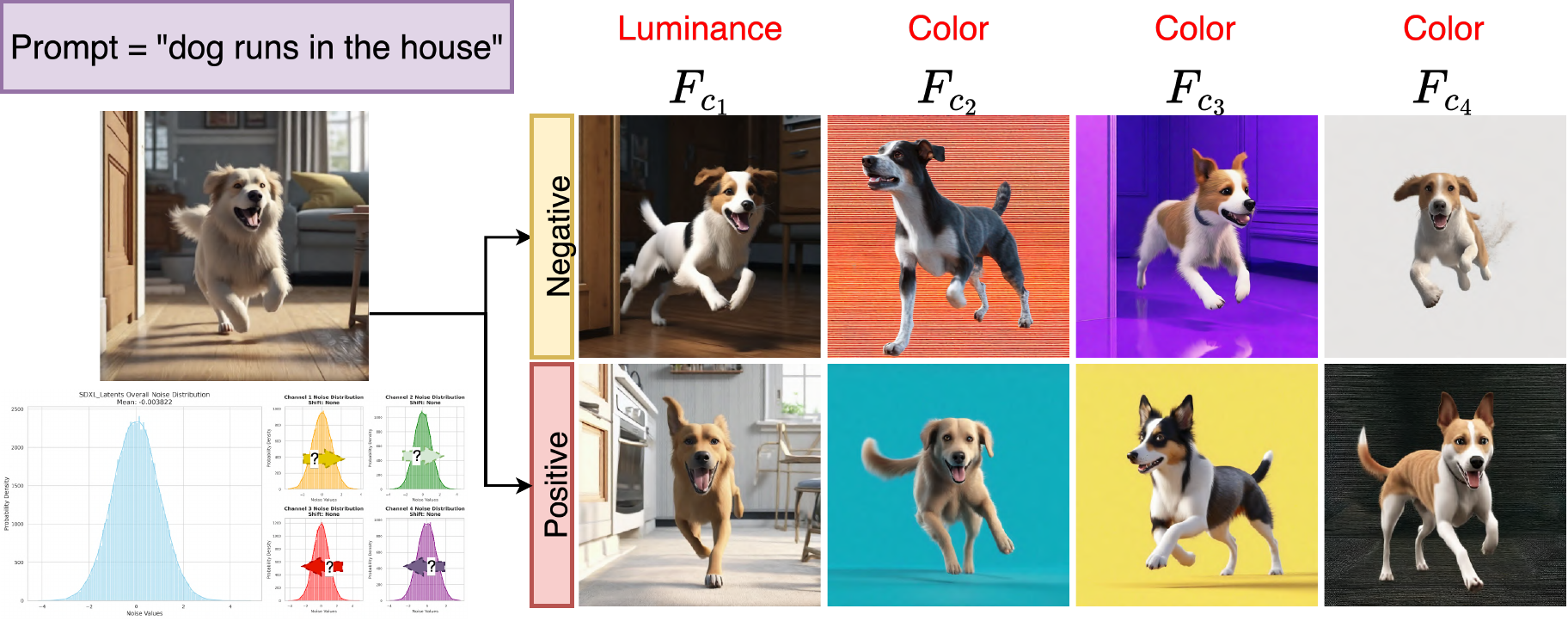}
    \caption{\textbf{Channel-wise sensitivity analysis via scaling the initial latent noise.}
    We scale a single latent channel $z_T^{(c)}$ ($c\in\{1,2,3,4\}$) by a constant factor $\alpha$,
    while keeping the prompt, random seed, and the other channels fixed.
    Scaling \emph{channel 1} consistently induces global brightness changes and alters the perceived illumination direction,
    whereas scaling channels 2--4 mainly affects chromatic attributes with limited impact on lighting.}  
    \label{fig:analysis_channel}
\end{figure}

% -------------------------------------------------------------------

\subsection{Analysis: Channel-wise Effects of Initial Noise}
We analyze how channel-wise perturbations of the initial latent noise influence illumination-related attributes in the generated images. Specifically, we conduct a controlled channel-wise perturbation on the initial latent noise $z_T \in \mathbb{R}^{H/8 \times W/8 \times 4}$.
For each channel $c \in \{1,2,3,4\}$, we apply a constant scaling while keeping all other channels unchanged:
\begin{equation}
    \hat{z}_T^{(c)} = \alpha \odot  z_T^{(c)},
\end{equation}
where $\alpha$ is a constant scaling factor.
We generate images using the same text prompt and random seed, varying only the perturbed channel, in order to isolate its effect.

Fig.~\ref{fig:analysis_channel} shows the qualitative results of this analysis.
We observe that perturbing \emph{channel 1} consistently induces global changes in brightness and alters the apparent illumination direction across the scene.
In contrast, perturbations applied to channels 2--4 primarily affect color tone and background hue, with minimal influence on lighting or shadow structure.

These observations indicate that \emph{channel 1} is strongly correlated with illumination-related factors in the latent space.
This channel-wise sensitivity analysis motivates our design choice to guide lighting by selectively manipulating \emph{channel 1} of the initial latent noise, as described in the following sections.

\subsection{Light Conditional Generation (LCG)}
\label{sec:lcg}
Generating lighting conditions requires modeling complex interactions between light, shadows, and reflections. To simplify this, we introduce Light Conditional Generation (LCG), where users specify light direction via a graphical interface by selecting a point or line indicating the light source. The system generates a light mask $m_l$ that defines the light’s origin and spread, creating smooth and natural lighting effects.

In our approach, we employ a linear gradient to transition from white to black based on the distance \(d\) of each pixel \((i, j)\) from the user-defined light source. Specifically, we use:
\begin{equation}
    m_l(i, j) = \max\!\Bigl(0,\;1 - \tfrac{d(i, j)}{r}\Bigr),
\end{equation}
where $r$ is a user-defined radius that controls the extent of the light’s influence. Within this radius, the mask value linearly decreases from 1 (white) at the light source to 0 (black) at $d(i, j) = r$. Pixels beyond this radius remain at 0, effectively modeling regions with negligible light.

This linear formulation offers a straightforward way to control the range and smoothness of the light gradation, making it more intuitive for users to specify how far the lighting should extend in the image. The resulting mask $m_l$ is applied to guide the lighting during image generation.

\subsection{Latent Space Light Guidance (LSLG)}
\label{sec:lculsm}
Building on the generated light mask $m_l$, we propose a Latent Space Light Guidance (LSLG) technique to guide lighting in Stable Diffusion’s latent space. The mask is applied to \emph{channel 1} of the initial noise $z_T$, modulating light intensity according to the user’s input. The transformation is defined as:
\begin{equation}
    \hat{z}^1_T = z^1_T \odot (1 + m_l)
\end{equation}
where $z_T^{1}$ represents the initial latent noise at timestep $T$ for channel 1, and $m_l$ scales light intensity across the scene.

Our experiments reveal that \emph{channel 1} encodes illumination-related information, making it the ideal target for light manipulation. Adjusting this channel allows for intuitive and precise control of lighting conditions directly in the generation process without additional training.

% -------------------------------------------------------------------------

\begin{figure*}[t]
    \centering
    \includegraphics[width=\linewidth]{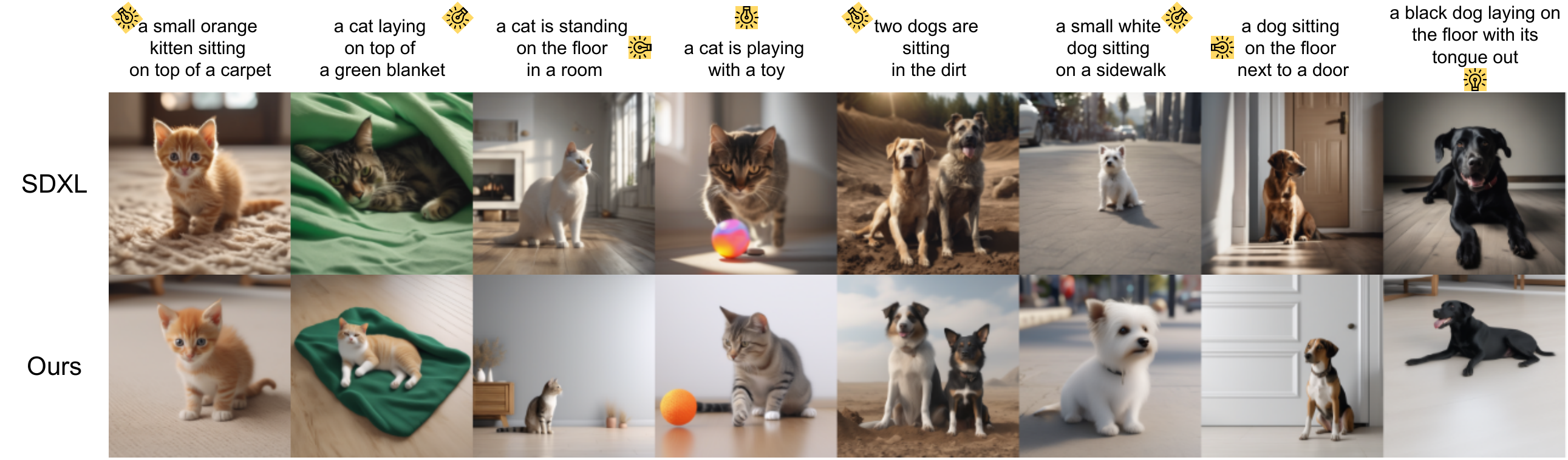}
    \caption{\textbf{Qualitative Results.} The existing model fails to control lighting conditions, often generating images with random or inconsistent lighting. In contrast, our approach effectively incorporates user-specified light direction and intensity, producing more natural and coherent lighting effects in the generated images.}
    \label{fig:qualitative}
\end{figure*}

% -------------------------------------------------------------------------

% -------------------------------------------------------------------------
\begin{table*}[t]
  \caption{Quantitative results comparing ours with the baseline. }
  \label{table:quantitative}
  \centering
  \setlength{\tabcolsep}{3mm}
  \resizebox{0.8\textwidth}{!}{% Adjust table to fit full text width
  \begin{tabular}{l|cccccc}
    \hline
    Method & FID $\downarrow$ & NIMA $\uparrow$ & CLIP-I $\uparrow$ & CLIP-T $\uparrow$ & Left $\uparrow$ & Right $\uparrow$ \\
    \hline
    \multicolumn{7}{c}{Cat} \\
    \hline
    SDXL  & \textbf{69.57}     & 5.44    & \textbf{0.671}     & 0.317     & 52.8\%  & 52.9\%  \\
    \textbf{Ours}  & 79.13  & \textbf{5.66} & 0.663   & \textbf{0.320}   & \textbf{79.0}\% & \textbf{77.3}\%  \\
    \hline
    \multicolumn{7}{c}{Dog} \\
    \hline
    SDXL  & \textbf{71.15}     & 5.67    & \textbf{0.618}     & 0.309     & 51.9\%  & 52.3\%  \\
    \textbf{Ours}  & 81.08  & \textbf{5.68} & 0.610   & \textbf{0.312}   & \textbf{77.3}\% & \textbf{76.7}\% \\
    \hline
  \end{tabular}
  }
\end{table*}
%-------------------------------------------------------------------------
\section{Experiments}
\label{sec:experiments}
\subsection{Experimental Setup}
\label{sec:setup}
We conduct experiments using the Stable Diffusion XL (SDXL)~\cite{podell2023sdxl} to generate images at a resolution of $1024 \times 1024$. For inference, we utilize the DDIMSampler with a guidance scale of 7.5 and 50 time steps. Since no prior work has addressed light-guided text-to-image generation, we use vanilla SDXL with prompt engineering as a baseline. Light-specific prompts such as ``light coming from the $light\_direction$'' are added to evaluate its ability to guide lighting.

\subsection{Dataset and Metrics}
\label{sec:data_metrics}
We use the Dog and Cat dataset~\cite{elson2007asirra}, containing 2,000 images evenly split between cats and dogs. Captions were generated using BLIP~\cite{li2022blip} to focus the comparison on light control rather than object generation accuracy. Limiting object categories enabled controlled experiments on light conditions.

To assess visual realism and aesthetics, we employ Fréchet Inception Distance (FID)~\cite{heusel2017gans} and Neural Image Assessment (NIMA)~\cite{talebi2018nima}. Text-image alignment is assessed using CLIP-I and CLIP-T~\cite{hessel2021clipscore}.
To evaluate light control, we propose light accuracy. First, we use YOLOv8~\cite{varghese2024yolov8} to detect the object and expand their bounding boxes by 1.25x to include surrounding areas. Within these regions, we apply a shadow detection model~\cite{cong2023sddnet} to analyze shadow directions. This metric assesses whether object shadows align correctly with the specified light direction, leveraging the principle that shadows extend opposite the light source.
For example, the shadow should extend to the right if the light is from the left. This metric thus provides a direct quantitative assessment of how effectively our model positions shadows according to the intended lighting.

\subsection{Qualitative Results}
\label{sec:qualitative_result}
Fig.~\ref{fig:qualitative} shows qualitative comparisons between the baseline and our method. While the baseline model generates high-quality images aligned with text prompts, it fails to account for light-specific directives, often placing light sources at random. In contrast, our model incorporates both text prompts and light direction, accurately control the specified light source. These results demonstrate our method's capability to understand and reflect light-shadow relationships.

\begin{table*}[t]
  \caption{Quantitative results with the ControlNet.}
  \label{table:quantitative_controlnet}
  \centering
  \setlength{\tabcolsep}{3mm}
  \resizebox{0.8\textwidth}{!}{% Adjust table to fit full text width
  \begin{tabular}{l|cccccc}
    \hline
    Method & FID $\downarrow$ & NIMA $\uparrow$ & CLIP-I $\uparrow$ & CLIP-T $\uparrow$ & Left $\uparrow$ & Right $\uparrow$ \\
    \hline
    \multicolumn{7}{c}{Cat} \\
    \hline
    ControlNet  & \textbf{71.56}     & 5.46    &  0.664  & 0.296     & 51.2\%  & 51.8\%  \\
    \textbf{Ours}  & 83.63  & \textbf{5.66} & \textbf{0.668}   & \textbf{0.315}   & \textbf{77.3}\% & \textbf{76.2}\% \\
    \hline
    \multicolumn{7}{c}{Dog} \\
    \hline
    ControlNet  & \textbf{76.63} & 5.56 & 0.606 & 0.283 & 51.6\% & 52.1\% \\
    \textbf{Ours} & 83.54 & \textbf{5.62} & \textbf{0.613} & \textbf{0.307} & \textbf{77.7}\% & \textbf{73.2}\% \\
    \hline
  \end{tabular}
  }
\end{table*}
% -------------------------------------------------------------------------

\subsection{Quantitative Results}
\label{sec:quantitative_result}
Table~\ref{table:quantitative} summarizes the scores for visual quality, text alignment, and light control accuracy. In terms of perceptual quality and text alignment, the two methods remain comparable, as reflected by similar NIMA and CLIP-based scores. However, FID increases with our method, which is consistent with its known sensitivity to global appearance changes such as brightness and illumination. Since LGTM steers lighting, the generated images deviate from the dataset’s marginal lighting distribution, even when perceptual quality is preserved. Thus, the higher FID reflects a controllability--distribution trade-off rather than a degradation in visual quality.

The critical difference lies in light control accuracy.
While the baseline model produces near-random shadow orientations (approximately 52\% for both left and right lighting), our method achieves substantially higher accuracy.
For example, under left-side lighting, LGTM correctly aligns shadows in 79.0\% (Cat) and 77.3\% (Dog) of the generated images, compared to 52.8\% and 51.9\% for the baseline.
Similarly, under right-side lighting, LGTM attains 77.3\% (Cat) and 76.7\% (Dog) accuracy, whereas the baseline remains close to chance level (52.9\% and 52.3\%).
These results demonstrate that LGTM provides reliable and consistent lateral lighting control without additional training, while maintaining high visual quality and text--image alignment.

% -------------------------------------------------------------------------
\begin{figure}[t]
    \centering
    \includegraphics[width=0.95\linewidth]{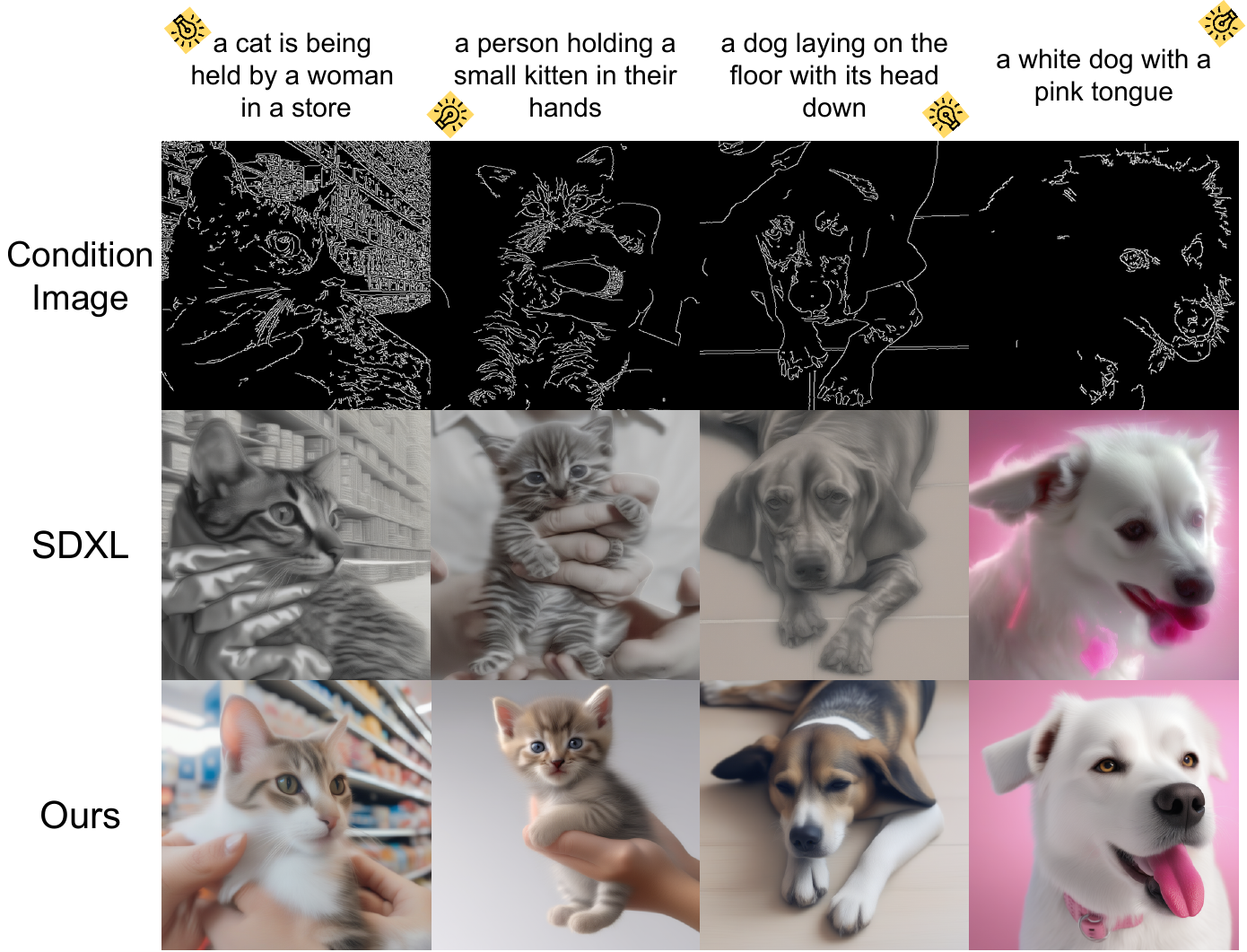}
    \caption{\textbf{Qualitative Results in Application.} Existing models integrated with ControlNet~\cite{zhang2023adding} control edges but fail to handle lighting, often producing inconsistent results. Our approach combines user-specified light direction and intensity with edge control, generating images with natural lighting and precise structure, demonstrating versatility in handling multiple controls.}
    \label{fig:control}
\end{figure}

\section{Application}
We demonstrate the LGTM's flexibility by integrating ControlNet~\cite{zhang2023adding}, enabling simultaneous control over text prompts, edges, and lighting.
LGTM only manipulates the initial latent noise, it can be plugged into a latent-based generation model straightforwardly and used jointly with structural conditions.

\subsection{Qualitative Comparison with Standard ControlNet}
Fig.~\ref{fig:control} illustrates a qualitative comparison between standard ControlNet and our extended method. While ControlNet successfully generates images conditioned on text prompts and canny edges, it fails to account for specified lighting directions. In contrast, our method incorporates all conditional information, including edge and light guidance, to generate images with more coherent lighting and shadow placement.

\subsection{Quantitative Comparison with Standard ControlNet}
As shown in Table~\ref{table:quantitative_controlnet}, our model surpasses the standard ControlNet in almost visual quality and text alignment metrics. Light control's accuracy highlights ControlNet's shortcomings in generating images aligned with light direction. In contrast, our method effectively integrates lighting conditions, ensuring accurate shadow placement and consistency.

\section{Limitations and Future Work}
\begin{figure}[t]
    \centering
    \includegraphics[width=0.9\linewidth]{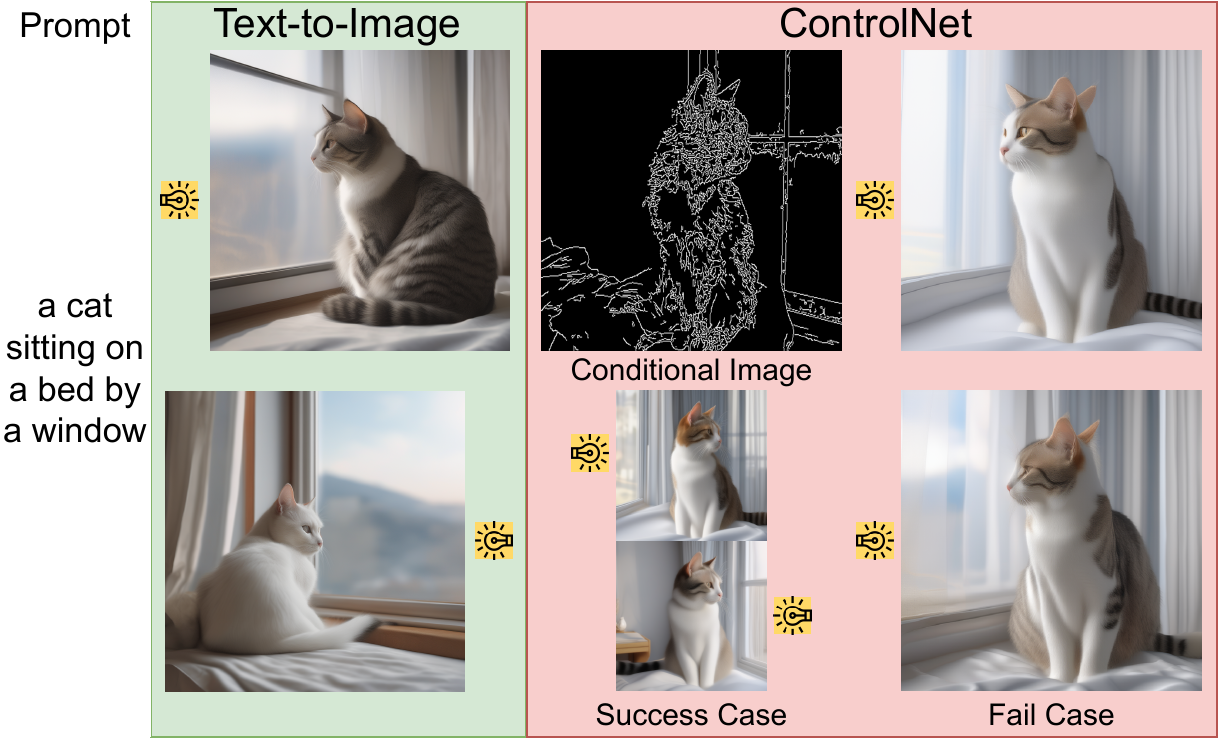}
    \caption{\textbf{Illustration of Light Alignment Behavior.} LGTM aligns generated subjects with the specified light direction, even overriding explicit constraints from ControlNet (e.g., canny-edge-based facial orientation). This can enhance realism but may also result in unintended subject orientation.}
    \label{fig:limit}
\end{figure}

As shown in Fig.~\ref{fig:limit}, the generated subjects tend to align their orientation with the direction of the light source.
This tendency persists even under ControlNet~\cite{zhang2023adding} constraints, indicating that the diffusion model prioritizes consistency with lighting cues over geometric orientation.
This limitation becomes apparent in scenarios that require independent control of lighting and subject pose.
As this work represents an initial step toward light-guided text-to-image generation, a deeper investigation into the interaction between lighting conditions and generative biases remains an important direction for future research.

% -------------------

% -------------------------------------------------------------------------
% -------------------------------------------------------------------------

\section{Conclusion}
This work is the first to explicitly explore the relationship between the VAE latent space in Stable Diffusion and light control, identifying \emph{channel 1} as key to intuitive and precise light manipulation.
Our method provides a user-friendly interface that integrates text prompts and light conditions, enabling seamless control of lighting during image generation. Furthermore, its adaptability allows integration with models like ControlNet, broadening its potential applications.
Extensive result show that LGTM effectively aligns generated images with both textual descriptions and user-specified light directions without additional training. This advancement highlights the practicality and versatility of our approach to dynamic, user-guided image generation.

\section*{Acknowledgements}
This work was supported by the BMBF Project Albatross (Grant 01IW24002). All compute was done thanks to the Pegasus cluster at DFKI.
% \clearpage

\bibliographystyle{unsrt}
\bibliography{ref}

% \bibitem{b1} G. Eason, B. Noble, and I. N. Sneddon, ``On certain integrals of Lipschitz-Hankel type involving products of Bessel functions,'' Phil. Trans. Roy. Soc. London, vol. A247, pp. 529--551, April 1955.
% \bibitem{b2} J. Clerk Maxwell, A Treatise on Electricity and Magnetism, 3rd ed., vol. 2. Oxford: Clarendon, 1892, pp.68--73.
% \bibitem{b3} I. S. Jacobs and C. P. Bean, ``Fine particles, thin films and exchange anisotropy,'' in Magnetism, vol. III, G. T. Rado and H. Suhl, Eds. New York: Academic, 1963, pp. 271--350.
% \bibitem{b4} K. Elissa, ``Title of paper if known,'' unpublished.
% \bibitem{b5} R. Nicole, ``Title of paper with only first word capitalized,'' J. Name Stand. Abbrev., in press.
% \bibitem{b6} Y. Yorozu, M. Hirano, K. Oka, and Y. Tagawa, ``Electron spectroscopy studies on magneto-optical media and plastic substrate interface,'' IEEE Transl. J. Magn. Japan, vol. 2, pp. 740--741, August 1987 [Digests 9th Annual Conf. Magnetics Japan, p. 301, 1982].
% \bibitem{b7} M. Young, The Technical Writer's Handbook. Mill Valley, CA: University Science, 1989.
% \end{thebibliography}
% \vspace{12pt}
% \color{red}
% IEEE conference templates contain guidance text for composing and formatting conference papers. Please ensure that all template text is removed from your conference paper prior to submission to the conference. Failure to remove the template text from your paper may result in your paper not being published.

\end{document}